\documentclass{article}

\usepackage{PRIMEarxiv}

\usepackage[utf8]{inputenc} 
\usepackage[T1]{fontenc}    
\usepackage{hyperref}       
\usepackage{url}            
\usepackage{booktabs}       
\usepackage{amsfonts}       
\usepackage{nicefrac}       
\usepackage{microtype}      
\usepackage{lipsum}
\usepackage{fancyhdr}       
\usepackage{graphicx}       
\usepackage{subfig}
\usepackage{cite}
\usepackage{times}
\usepackage{epsfig}
\usepackage{graphicx}
\usepackage{amsmath}
\usepackage{amssymb}
\usepackage{float}
\title{Dynamic Clone Transformer for Efficient Convolutional Neural Netwoks
}

\author{
  LongQing Ye
}

\begin{document}
\maketitle

\begin{abstract}
Convolutional networks (ConvNets) have shown impressive capability to solve various vision tasks. Nevertheless, the trade-off between performance and efficiency is still a challenge for a feasible model deployment on resource-constrained platforms. In this paper, we introduce a novel concept termed multi-path fully connected pattern (MPFC) to rethink the interdependencies of topology pattern, accuracy and efficiency for ConvNets. Inspired by MPFC, we further propose a dual-branch module named dynamic clone transformer (DCT) where one branch generates multiple replicas from inputs and another branch reforms those clones through a series of difference vectors conditional on inputs itself to produce more variants. This operation allows the self-expansion of channel-wise information in a data-driven way with little computational cost while providing sufficient learning capacity, which is a potential unit to replace computationally expensive pointwise convolution as an expansion layer in the bottleneck structure.
\end{abstract}

\section{Introduction}

Convolutional layer has been widely applied in ConvNets as a basic building block to replace fully connected layer due to its  high efficiency and powerful expressivity. Deeper (more layers) or wider (more channels) networks \cite{alex,wide,vgg} have been developed for higher performance, leading to larger model capacity and more computational overhead. As a result, tremendous resource consumption severely impedes the deployment of those large models on resource-constrained hardware environments for various vision applications. Therefore, how to trade off accuracy and efficiency becomes an important research topic in the field of model design.

\begin{figure}[ht]
\centering
\includegraphics[width=0.6\linewidth]{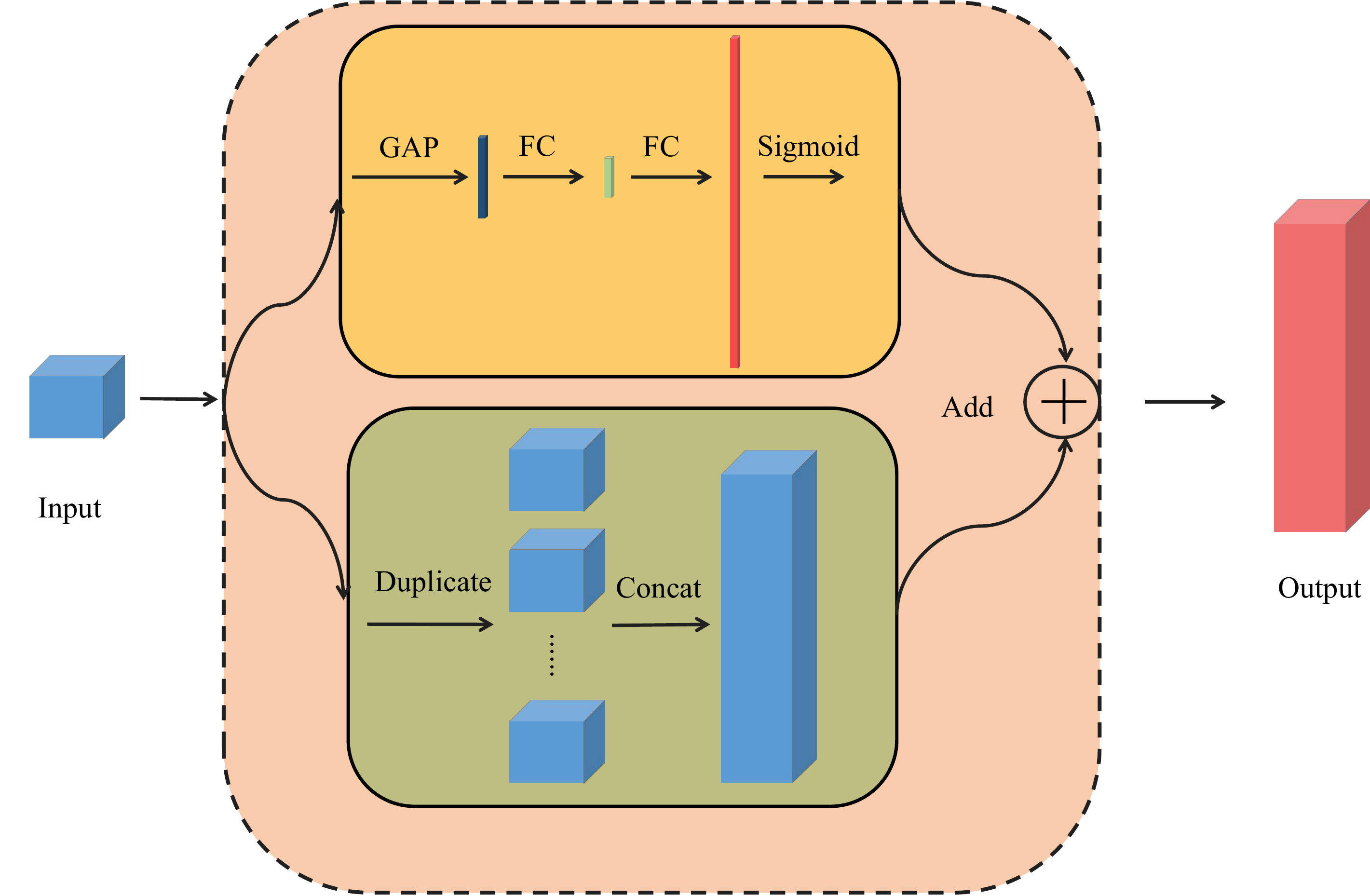}

\caption{The schema of dynamic clone transformer. GAP: global average pooling, FC: fully connected layer, Concat: concatenate feature maps along channel dimension.}
\label{module}
\end{figure}

In this paper, we propose a novel dual-branch module termed dynamic clone transformer (DCT). As depicted in Figure \ref{module}, DCT adopts parallel structure where one branch performs the replication operation to obtain more replicas of input channels and another branch adaptively generates global information to compensate another branch. DCT module can achieve the self-expansion of feature maps without much overhead while obtaining rich representational information in a data-driven way.

We notice that current bottleneck structure and variants \cite{resnet,mobilenetv2,ghostnet,rethinking} at least contain two steps with critical consumption in terms of computational cost and parameter count: contraction and expansion transformations using pointwise convolutional layers. In this paper, we make an attempt to exploit DCT module to reduce the computational cost of expansion process in the bottleneck building block while maintaining sufficient parameter capacity for a good model performance.

Specifically, using the DCT module to replace pointwise convolution as the expansion layer, we develop a novel efficient bottleneck variant named DCT-bottleneck. In the channel perspective, the DCT-bottleneck follows contraction and expansion design pattern while requiring less computational cost at the expansion stage compared to other bottleneck variants.

\begin{figure*}[ht]
\centering
\subfloat[]{\label{n1}\includegraphics[scale=0.4]{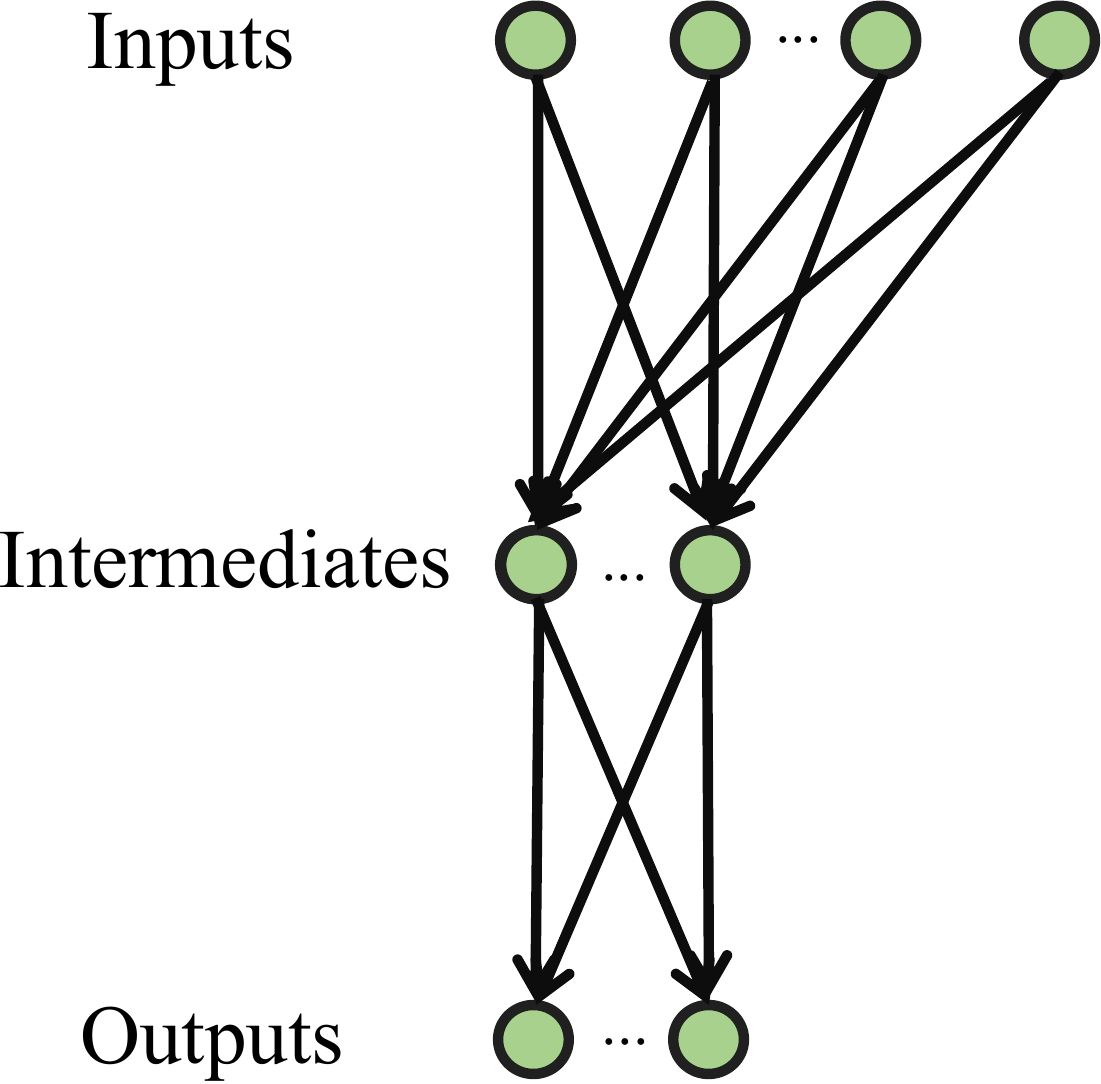}}
\hspace{10mm}
\subfloat[]{\label{n2}\includegraphics[scale=0.4]{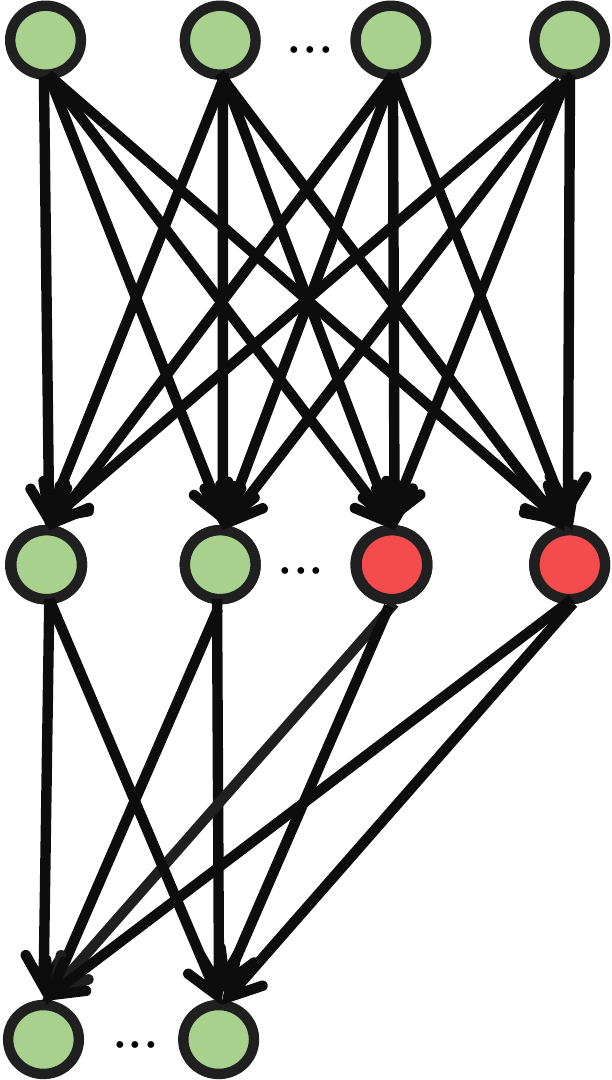}}
\hspace{10mm}
\subfloat[]{\label{n3}\includegraphics[scale=0.4]{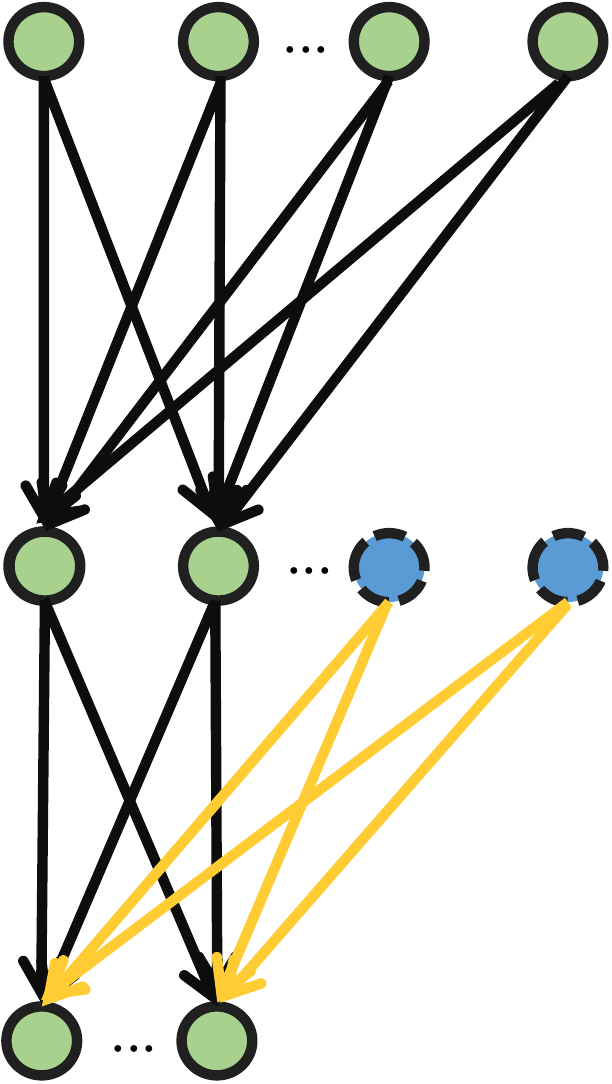}}
\hspace{10mm}
\subfloat[]{\label{n4}\includegraphics[scale=0.4]{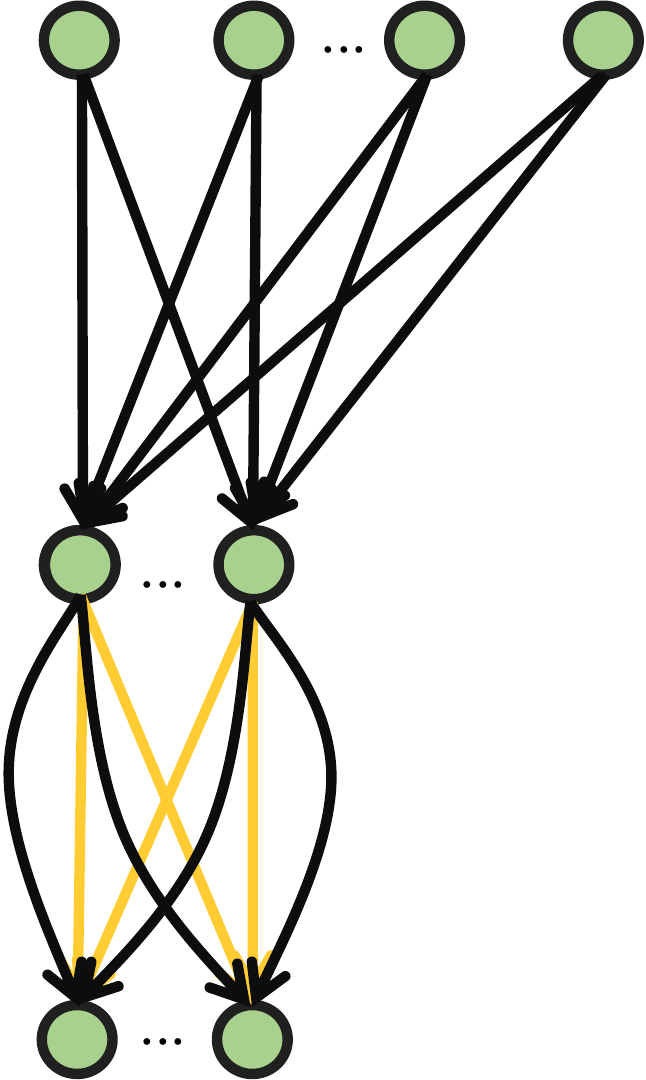}}

\caption{Topology structure of different networks. a) Original Network; b) Increase the width of network; c) Duplicate nodes to increase the width of network; d) Multi-path fully connected structure (the second layer of network) where there are multiple paths between every pair of nodes from inputs and outputs, respectively. Note that the second layers of (c) and (d) are equivalent.}
\label{cdg}
\end{figure*}

\section{Related Work}

Recent efforts to build compact ConvNets can be roughly categorized into efficient model design and model compression. Efficient model design aims to explore highly efficient convolution operators, building blocks and augmented modules to construct models with both high performance and efficiency. Model compression utilizes pruning strategy\cite{prun1,channelprun}, low-bit representation \cite{deep,binarized} and other approaches \cite{distilling,low-rank} to shrink a large pre-trained model, which is usually used as a post-processing way.

\subsection{Efficient Model Design}

Composition of low-rank ($1\times3$, $3\times1$) filters \cite{low-rank} or sparse filters \cite{shufflenet} are proposed to approximate dense convolution filter. Those work indicates that less redundant filters can bring a great reduction of FLOPs and parameters while maintaining the performance of models. Group convolution has been viewed as a standard operator in modern compact models \cite{resnext,shufflenet}. Depth-wise convolution is an extreme form of group convolution, in which each channel presents a group. ShuffleNetV1 \cite{shufflenet} uses group convolution to replace $1\times1$ convolution and further introduces the operation of channel shuffle to improve cross-group information communication. MobilenetV1 \cite{mobilenetv1} utilizes depth-wise convolution and pointwise convolution to construct a lightweight model for mobile platforms. As for the design pattern of compact building blocks, ResNet\cite{resnet} adopts a bottleneck structure to alleviate the burden of heavy computation for channel expansion. MobileNetV2 \cite{mobilenetv2} proposes inverted residual block and further improves the bottleneck structure.

\subsection{Model Compression}

Comlementary to compact model design, model compression is another approach to further shrink pre-trained models. Network prunning \cite{channelprun} removes redundant and non-informative connections or channels. Model quantization \cite{quantized} aims to represent stored weights at a low cost for model compression and calculation acceleration. Knowledge distillation \cite{distilling} transfers refined knowledge from "teacher network" into "student network", simplifying the process of suppressing redundant information. In addition, efficient convolution algorithms like FFT \cite{fft} and winograd \cite{fast} are explored to speed up the implement of convolutional layer without any modification of network design.

\section{Method} \label{approach}
In this section, we first introduce a novel topology pattern: multi-path fully connected (MPFC) structure. Inspired by the MPFC structure, we further propose the dynamic clone transformer as a lightweight expansion unit to improve the efficiency of bottleneck structure.

\subsection{Multi-path Fully Connected Structure} \label{MPFC}

Consider a network with $m$ input nodes, $n$ output nodes, and $l$ intermediate nodes, as illustrated in Figure \ref{cdg}\subref{n1}. We utilize the node graph to analyze the connectivity pattern of the network. Figure \ref{cdg}\subref{n1} can be used to describe the dependency relationship of stacking fully connected layers or channel-wise dependency relationship of regular convolutional layers. For simplicity, we regard the network in Figure \ref{cdg}\subref{n1} as three fully connected layers where each node represents a neuron. Thus, the overall computational cost of network is $l(m+n)$. As we increase $l$ to $l+\Delta l$ for larger model capacity, as illustrated in Figure \ref{cdg}\subref{n2}, additional overhead $\Delta l(m+n)$ mainly coming from the current and previous layers is considerable if $m$ and $n$ are both large enough. To reduce extra burden, as shown in Figure \ref{cdg}\subref{n3}, a simple solution is to duplicate intermediate nodes such that the additional overhead finally turns into $\Delta l \times n$. Therefore, the replication operation can effectively increase the capacity of the network in a moderate way by confining the extra overhead within only the current layer.

It is seen that the replication operation is concerned with a more general connectivity pattern: multi-path fully connected (MPFC) structure as illustrated as Figure \ref{cdg}\subref{n4}. Essentially, the proposed MPFC structure is an extension of conventional fully connected structure which allows multiple paths between every pair of nodes (channels or neurons) from inputs and outputs, respectively.

Formally, every output node of FC pattern can be written as:
\begin{equation}
{y}_{i}=\sum\limits_{j=1}^{s}{{w}_{ij}}{x}_{j} \quad i=1,2,3,...
\end{equation}
where $x_{j}$ represents $j$-th input node, $s$ denotes the number of input nodes. $w_{ij}$ represents the weight or the connections between input and output nodes.

Correspondingly, every output node of MPFC pattern can be formulated as

\begin{equation}
{y}_{i}=\sum\limits_{k=1}^{p}{\sum\limits_{j=1}^{s}{{w}_{ijk}}}{x}_{j} \quad i=1,2,3,...
\end{equation}

\noindent where $p$ denotes the number of paths between every pair of nodes from inputs to outputs. It is seen that FC structure can be regarded as a special case of MPFC when $p=1$. Theoretically, MPFC structure ($p>1$) has a more powerful representation ability than FC structure ($p=1$) given the same number of input and output nodes, since the former owns larger parameter capacity. In this paper, we postulate that low-dimension MPFC pattern can approximate the high-dimension FC pattern in terms of representation ability if configured properly. Motivated by this, we attempt to utilize the MPFC pattern to improve the efficiency of ConvNets, which will be discussed in section \ref{dynamic clone transformer}.

\subsection{Dynamic Clone Transformer} \label{dynamic clone transformer}
For a regular convolutional layer, each output channel is actually connected to all input channels from the perspective of channel dimension, following a FC pattern. Given a regular convolutional layer with $n$ convolutional filters $
F=\left \{{f}_{1}, {f}_{2},...., {f}_{n}
\right \}\in {\mathbb{R} }^{ d\times d\times c\times n}
$, where $d \times d$ is the size of filters and $c$ is the number of input channels, the convolution operation can be formulated as
\begin{equation}
 Y=  X\otimes  F
\label{conv1}
\end{equation}
where $X$ and $Y$ are input data and output features of this layer, $\otimes$ denotes the convolution operator.

We propose the dynamic clone transformer (DCT) module aiming to improve the efficiency of regular convolutional layers. Referring to equation \ref{conv1}, our approach can be formulated as

\begin{equation}
{X}^{'}=X\otimes {F}^{'}
\label{conv2}
\end{equation}

\begin{equation}
Y= D\left ({X}^{'}, p
\right ) +\Psi \left ({X}^{'}, \theta
\right )
\label{myconv}
\end{equation}

\noindent where ${F}^{'}$ represents $n^{'}$ convolutional filters ($n^{'} = \frac{n}{p}$). The $D$ function performs the replication operation where $p$ denotes the replication ratio for channel expansion. The $\Psi$ function is used to generates dynamic coefficients.

We first project the input data $X\in {\mathbb{R} }^{ d\times d\times c\times n}$ into ${X}^{'}\in {\mathbb{R} }^{ d\times d\times c\times n^{'}}$ with a small amount of filters as shown in equation \ref{conv2}. Then we use DCT module to perform expansion operation for more channels, corresponding to equation \ref{myconv}. In equation \ref{myconv}, the feature map ${X}^{'}$ is duplicated using the $D$ function. The second term $\Psi$ function in equation \ref{myconv} is employed as a recalibrating component to compensate the first term and further enhance the representation ability of model. Note that $\theta$ denotes all trainable parameters for the recalibrating component. More details will be discussed in \ref{implement}.

\subsection{Implementations} \label{implement}

The whole pipeline of dynamic clone transformer is illustrated as Figure \ref{module}. There are two main components: replicator and recalibrator.\\

\noindent\textbf{Replicator} We utilize a low-dimension MPFC pattern to modify the high-dimension FC pattern of channels in a regular convolutional layer, which can be implemented with a simple replication operation. Therefore, the replicator is used to duplicate the "squeezed" feature map, corresponding to the $D$ function in equation \ref{myconv}. The replication operation can effectively reduce the cost of expansion process yet also bring following side effects:
\begin{itemize}
\item[1.]Drastic decrease of parameters amount can impair the representation ability of models.

\item[2.]Indiscriminate duplication may introduce more useless information. For example, dead and weak channels are also duplicated, most of which may have little value yet still occupy computational resource.
\end{itemize}

To avoid performance degradation without introducing much computational cost, the replicator requires more supplementary parameters.\\

\noindent\textbf{Recalibrator} In response to above issues, the recalibrator servers as a complementary component to ensure sufficient capacity for the whole building block and generates dynamic global context for all clones from the replicator. In this work, the recalibrator is implemented by a tiny network similar to SE module \cite{se}. It is formulated as follows:

\begin{equation}
\Psi \left (X,\theta
\right )=\sigma \left ({h}_{2}\left ({h}_{1}\left (GAP\left (X
\right ),{\theta }_{1},r
\right ), {\theta }_{2}
\right )
\right )
\end{equation}

\noindent where $\sigma$ and $h$ denote the sigmoid function and fully connected layer, respectively. GAP represents the operation of global average pooling. Reduce ratio $r$ is a hyper-parameter to agjust the capacity of the recalibrator network. The output of recalibrator is merged into "feature clones" from the replicator via summation operation. This procedure forces the recalibrator to learn example-independent patterns to modify original feature, and reactivate those dead or weak neurons in the replicator.

\section{Network Design}

\subsection{Separable Convolution}

The separable convolution applied in MobileNetV1\cite{mobilenetv1} and Xception\cite{xception} has been proved very efficient, which factorizes a regular convolution into a spatial depth-wise convolution and a pointwise ($1\times1$) convolution. The depth-wise convolution applies a single convolutional filter independently for each channel of the inputs, extracting local spatial features. The pointwise convolution is used for information fusion for all input channels. This factorization brings drastic reduction in computation and model size. However, the overhead of separable convolution is dominated by the pointwise convolution.

Suppose a separable convolutional layer takes $M$ feature maps with a spatial size of $D_f \times D_f$ as inputs, and outputs $N$ feature maps of the same spatial size with appropriate padding. $M$ and $N$ denotes the number of the input and output channels, respectively. The filter size is $D_k \times D_k$ and the stride is 1 for the depth-wise convolutional layer. The computational cost of separable convolution $C_0$ is \newline
\begin{equation}
\begin{aligned}
D_f\times D_f \times D_k \times D_k \times M + D_f\times D_f \times M \times N
\end{aligned}
\end{equation}
\noindent where $M$ and $N$ are much larger than $D_k$ especially in the deep layers of the ConvNets. As a result, the $M\times N$ term becomes the denominator of computational cost $C_0$.

\subsection{DCT-based Bottleneck}

\begin{figure*}[ht]
\centering
\subfloat[]{\label{block1}\includegraphics[width=0.3\linewidth]{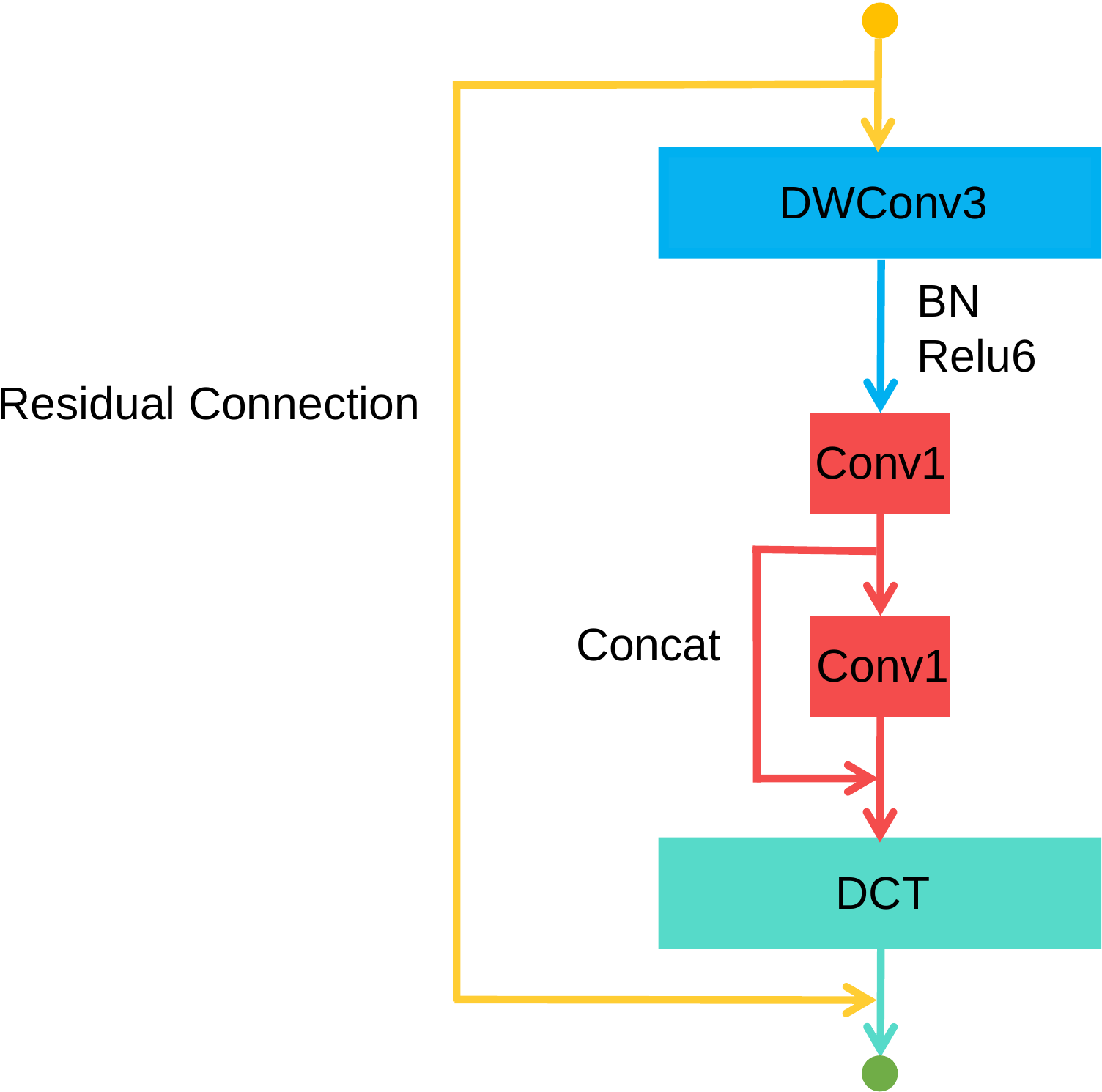}}
\hspace{20mm}
\subfloat[]{\label{block2}\includegraphics[width=0.3\linewidth]{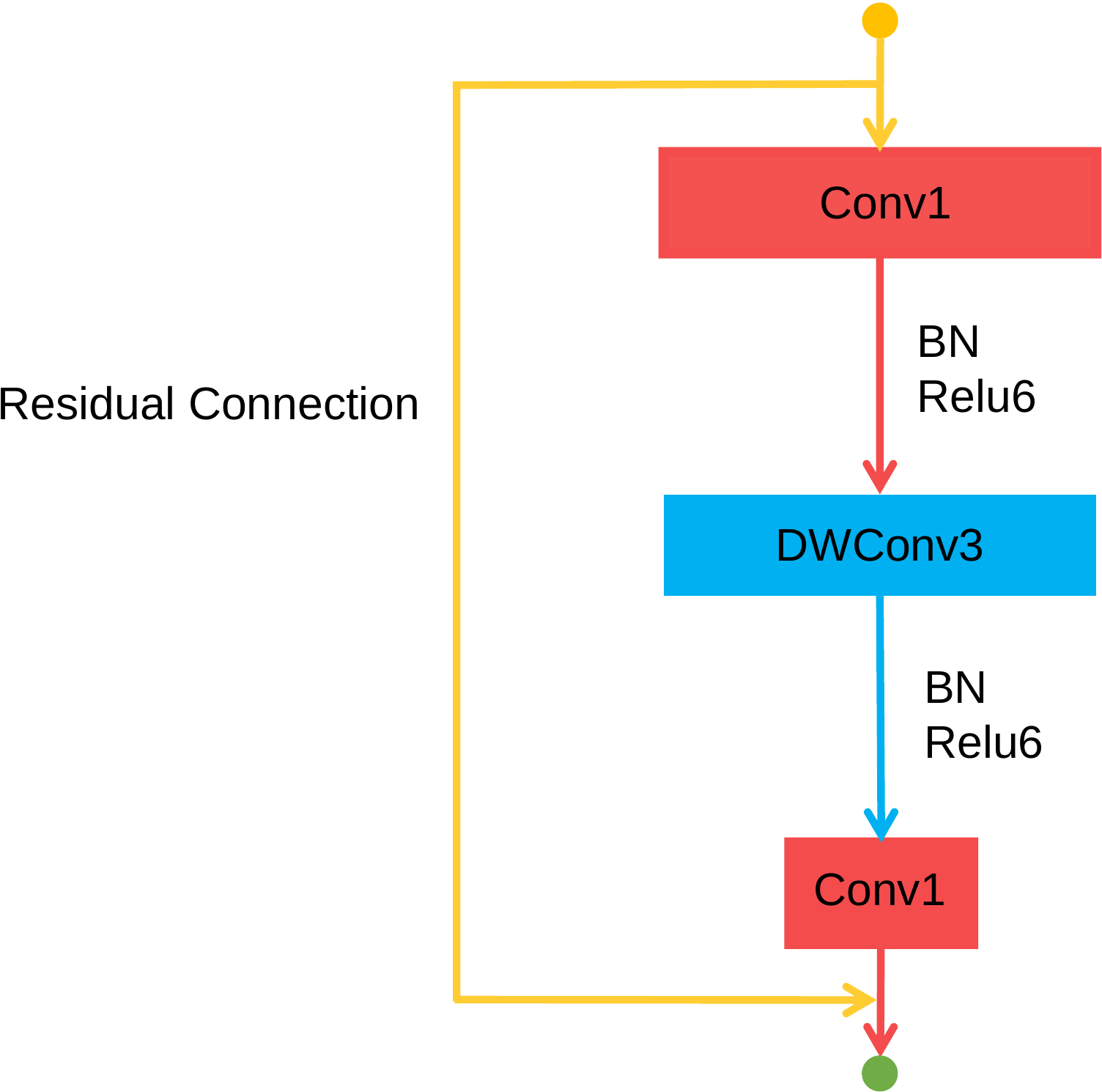}}

\caption{Different efficient building blocks based on separable convolution. a) Our DCT-based bottleneck structure; b) Inverted residual block in MobileNetV2.}
\label{block}
\end{figure*}

Bottleneck structure or other variants is widely used to squeeze the remaining room of efficiency for separable convolution layers by decoupling a high-dimensional transformation into two steps regardless of the order: contraction and expansion. In this paper, we combine the proposed dynamic clone transformer to modify the separable convolution, and further develop the efficient building block for our models.

As illustrated in Figure \ref{block}\subref{block1}, our building block named DCT-bottleneck follows a contraction-expansion design pattern in the channel dimension, consisting of a $3\times 3$ depth-wise convolutional layer, two pointwise ($1\times 1$) convolutional layers and the proposed dynamic clone transformer module. From the perspective of channel dimension, the first pointwise convolution is used to compress the input feature maps into a low-dimension representation. The second pointwise convolution is used to obtain a linear combination of input feature maps. For the first channel expansion and feature reuse, we concatenate the feature maps generated by two pointwise convolutional layers, respectively. At last, the obtained feature maps are duplicated through the proposed dynamic clone transformer in an adaptive way, achieving the second expansion for channels. It is seen that the computational cost of expansion process is much less than that of contraction process for DCT-based bottleneck. The detailed structure named DCT-based bottleneck for our building block is shown in Table \ref{inout}.

\begin{table}[ht]
\centering

\begin{tabular}{|l|c|c|r}
\hline
 Input & Operator & Output \\
\hline
\hline

${D_f}^{2}\times M$  & $3\times 3$ dwconv & ${D_f}^{2}\times M$ \\
${D_f}^{2}\times M$ & $1\times 1$ conv2d & ${D_f}^{2}\times \frac{N}{2p}$\\

${D_f}^{2}\times \frac{N}{2p} $ & $1\times 1$ conv2d & ${D_f}^{2}\times \frac{N}{2p}$\\

${D_f}^{2}\times \frac{N}{p}$ & DCT Module & ${D_f}^{2}\times N$ \\
\hline
\end{tabular}
\caption{The structure of DCT-based bottleneck.}
\label{inout}
\end{table}

\begin{table*}[h]
\begin{center}
\begin{tabular}{ |l|c|c|c|c| }
 \hline
 Input & Operator & \#repeat  & \#out & Stride \\
\hline
\hline

 $32^{2}\times3$ & Conv2d $3\times3$ & - & 32 & 1 \\ 

 $32^{2}\times32$ & bottleneck $3\times3$ & 5 & 64 & 1\\

 $32^{2}\times64$ & bottleneck $3\times3$ & 1 & 128 & 2\\

 $16^{2}\times128$ & bottleneck $3\times3$ & 5 & 128 & 1\\

 $16^{2}\times128$ & bottleneck $3\times3$ & 1 & 256 & 2\\

 $8^{2}\times256$ & bottleneck $3\times3$ & 6 & 256 & 1\\

$8^{2}\times256$ & Pool $8\times8$ & - & - & -\\

$1^{2}\times256$ & FC  & - & num\_classes & -\\
\hline
\end{tabular}
\caption{Network Architecture. The bottleneck denotes DCT-based bottleneck. \#repeat denotes the repetition number of DCT-based bottleneck building blocks.}
\label{net}
\end{center}
\end{table*}



As the computation complexity of DCT module is negligible, the overall computational cost of DCT-based bottleneck is calculated as
\begin{equation}
\begin{aligned}
 C_1 &={D_f}^{2} \times {D_k}^{2} \times M + {D_f}^{2} \times \frac{ MN}{2p} + {D_f}^{2} \times \frac{N^{2}}{4p^{2}} \\
&\approx {D_f}^{2} \times \left ( \frac{MN}{2p} + \frac{N^2}{4p^2}\right)\\
\end{aligned}
\end{equation}

Assume $M=N$ and $p\geq 2$, thus we have $C_{1}\leq\frac{5}{16}C_0$. Obviously, DCT-based bottleneck shows better efficiency in terms of computational cost compared to the separable convolution layer.

\subsection{Architecture Design}
A new convolutional network, named DyClotNet, is constructed by stacking the DCT-based bottleneck building blocks. The overall architecture of the network is shown in Table \ref{net}. In order to control the scale of models, three hyper-parameters: width multiplier $\alpha$, expansion ratio $p$, reduction ratio $r$ are introduced to configure the DyClotNets. The width multiplier $\alpha$ is used to thin a network uniformly by scaling the number of channels at each layer. DyClotNet with width multiplier $\alpha$ is noted as DyClotNet-$\alpha$. Expansion ratio $p$ and reduction ratio $r$ are two crucial and complementary factors to strike a balance between efficiency and performance for building a compact and powerful instance of DCT-based bottleneck.

\section{Conclusion}

In this work, we present the DCT module, a lightweight architectural unit which can perform adaptive self-expansion for channels. Utilizing the DCT module to replace computationally expensive pointwise convolution as an expansion layer, we further develop an efficient and powerful building block termed DCT-based bottleneck.

\bibliographystyle{unsrt}
\bibliography{references}

\begin{thebibliography}{10}

\bibitem{alex}
Alex Krizhevsky, Ilya Sutskever, and Geoffrey~E Hinton.
\newblock Imagenet classification with deep convolutional neural networks.
\newblock In {\em Advances in neural information processing systems}, pages
  1097--1105, 2012.

\bibitem{wide}
Sergey Zagoruyko and Nikos Komodakis.
\newblock Wide residual networks.
\newblock {\em arXiv preprint arXiv:1605.07146}, 2016.

\bibitem{vgg}
Karen Simonyan and Andrew Zisserman.
\newblock Very deep convolutional networks for large-scale image recognition.
\newblock {\em arXiv preprint arXiv:1409.1556}, 2014.

\bibitem{resnet}
Kaiming He, Xiangyu Zhang, Shaoqing Ren, and Jian Sun.
\newblock Deep residual learning for image recognition.
\newblock In {\em Proceedings of the IEEE conference on computer vision and
  pattern recognition}, pages 770--778, 2016.

\bibitem{mobilenetv2}
Mark Sandler, Andrew Howard, Menglong Zhu, Andrey Zhmoginov, and Liang-Chieh
  Chen.
\newblock Mobilenetv2: Inverted residuals and linear bottlenecks.
\newblock In {\em Proceedings of the IEEE conference on computer vision and
  pattern recognition}, pages 4510--4520, 2018.

\bibitem{ghostnet}
Kai Han, Yunhe Wang, Qi~Tian, Jianyuan Guo, Chunjing Xu, and Chang Xu.
\newblock Ghostnet: More features from cheap operations.
\newblock In {\em Proceedings of the IEEE/CVF Conference on Computer Vision and
  Pattern Recognition}, pages 1580--1589, 2020.

\bibitem{rethinking}
Zhou Daquan, Qibin Hou, Yunpeng Chen, Jiashi Feng, and Shuicheng Yan.
\newblock Rethinking bottleneck structure for efficient mobile network design.
\newblock {\em arXiv preprint arXiv:2007.02269}, 2020.

\bibitem{prun1}
Song Han, Jeff Pool, John Tran, and William~J Dally.
\newblock Learning both weights and connections for efficient neural networks.
\newblock {\em arXiv preprint arXiv:1506.02626}, 2015.

\bibitem{channelprun}
Yihui He, Xiangyu Zhang, and Jian Sun.
\newblock Channel pruning for accelerating very deep neural networks.
\newblock In {\em Proceedings of the IEEE International Conference on Computer
  Vision}, pages 1389--1397, 2017.

\bibitem{deep}
Suyog Gupta, Ankur Agrawal, Kailash Gopalakrishnan, and Pritish Narayanan.
\newblock Deep learning with limited numerical precision.
\newblock In {\em International conference on machine learning}, pages
  1737--1746. PMLR, 2015.

\bibitem{binarized}
Itay Hubara, Matthieu Courbariaux, Daniel Soudry, Ran El-Yaniv, and Yoshua
  Bengio.
\newblock Binarized neural networks.
\newblock In {\em Proceedings of the 30th international conference on neural
  information processing systems}, pages 4114--4122. Citeseer, 2016.

\bibitem{distilling}
Geoffrey Hinton, Oriol Vinyals, and Jeff Dean.
\newblock Distilling the knowledge in a neural network.
\newblock {\em arXiv preprint arXiv:1503.02531}, 2015.

\bibitem{low-rank}
Max Jaderberg, Andrea Vedaldi, and Andrew Zisserman.
\newblock Speeding up convolutional neural networks with low rank expansions.
\newblock {\em arXiv preprint arXiv:1405.3866}, 2014.

\bibitem{shufflenet}
Xiangyu Zhang, Xinyu Zhou, Mengxiao Lin, and Jian Sun.
\newblock Shufflenet: An extremely efficient convolutional neural network for
  mobile devices.
\newblock In {\em Proceedings of the IEEE conference on computer vision and
  pattern recognition}, pages 6848--6856, 2018.

\bibitem{resnext}
Saining Xie, Ross Girshick, Piotr Doll{\'a}r, Zhuowen Tu, and Kaiming He.
\newblock Aggregated residual transformations for deep neural networks.
\newblock In {\em Proceedings of the IEEE conference on computer vision and
  pattern recognition}, pages 1492--1500, 2017.

\bibitem{mobilenetv1}
Andrew~G Howard, Menglong Zhu, Bo~Chen, Dmitry Kalenichenko, Weijun Wang,
  Tobias Weyand, Marco Andreetto, and Hartwig Adam.
\newblock Mobilenets: Efficient convolutional neural networks for mobile vision
  applications.
\newblock {\em arXiv preprint arXiv:1704.04861}, 2017.

\bibitem{quantized}
Jiaxiang Wu, Cong Leng, Yuhang Wang, Qinghao Hu, and Jian Cheng.
\newblock Quantized convolutional neural networks for mobile devices.
\newblock In {\em Proceedings of the IEEE Conference on Computer Vision and
  Pattern Recognition}, pages 4820--4828, 2016.

\bibitem{fft}
Michael Mathieu, Mikael Henaff, and Yann LeCun.
\newblock Fast training of convolutional networks through ffts.
\newblock {\em arXiv preprint arXiv:1312.5851}, 2013.

\bibitem{fast}
Andrew Lavin and Scott Gray.
\newblock Fast algorithms for convolutional neural networks.
\newblock In {\em Proceedings of the IEEE Conference on Computer Vision and
  Pattern Recognition}, pages 4013--4021, 2016.

\bibitem{se}
Jie Hu, Li~Shen, and Gang Sun.
\newblock Squeeze-and-excitation networks.
\newblock In {\em Proceedings of the IEEE conference on computer vision and
  pattern recognition}, pages 7132--7141, 2018.

\bibitem{xception}
Fran{\c{c}}ois Chollet.
\newblock Xception: Deep learning with depthwise separable convolutions.
\newblock In {\em Proceedings of the IEEE conference on computer vision and
  pattern recognition}, pages 1251--1258, 2017.

\end{thebibliography}

\end{document}